\documentclass[10pt,letterpaper]{article}
\usepackage[top=0.85in,left=2.75in,footskip=0.75in]{geometry}
\usepackage[T1]{fontenc}
\usepackage{lmodern}

\usepackage{amsmath,amssymb}

\usepackage{changepage}

\usepackage{textcomp,marvosym}

\usepackage{cite}

\usepackage{nameref,hyperref}

\usepackage[right]{lineno}

\usepackage[nopatch=eqnum]{microtype}
\DisableLigatures[f]{encoding = *, family = * }

\usepackage[table]{xcolor}

\usepackage{array}

\newcolumntype{+}{!{\vrule width 2pt}}

\newlength\savedwidth

\raggedright
\usepackage[aboveskip=1pt,labelfont=bf,labelsep=period,justification=raggedright,singlelinecheck=off]{caption}

\makeatletter
\renewcommand{\@biblabel}[1]{\quad#1.}
\makeatother

\usepackage{lastpage,fancyhdr,graphicx}
\usepackage{epstopdf}
\fancyheadoffset[L]{2.25in}
\fancyfootoffset[L]{2.25in}
\usepackage{amsmath,amssymb,amsfonts}
\usepackage{algorithmic}
\usepackage{graphicx}
\usepackage{textcomp}
\usepackage{multirow}
\usepackage{colortbl} 
\definecolor{mygray}{gray}{0.75}
\newcommand{\grayline}{\arrayrulecolor{mygray}\hline\arrayrulecolor{black}}
\usepackage{hyperref}

\begin{document}
\vspace*{0.2in}

\begin{flushleft}
{\Large
\textbf\newline{Fully automated workflow for \textcolor{black}{designing} patient-specific orthopaedic implants: application to total knee arthroplasty} 
}
\newline
\\
Aziliz Guezou-Philippe\textsuperscript{1,2},
Arnaud Clavé\textsuperscript{1,3},
Ehouarn Maguet\textsuperscript{1},
Ludivine Maintier\textsuperscript{1},
Charles Garraud\textsuperscript{1,4},
Jean-Rassaire Fouefack\textsuperscript{1},
Valérie Burdin\textsuperscript{1,2},
Eric Stindel\textsuperscript{1,4,5},
Guillaume Dardenne\textsuperscript{1}
\\
\bigskip
\textsuperscript{1} LaTIM - INSERM UMR 1101, Brest, France \\
\textsuperscript{2} IMT Atlantique, Brest, France \\
\textsuperscript{3} Clinique Saint George, Nice, France \\
\textsuperscript{4} Brest University Hospital, Brest, France \\
\textsuperscript{5} University of Western Brittany, Brest, France \\
\bigskip

\end{flushleft}

\section*{Abstract}

\textbf{Background} Osteoarthritis affects about 528 million people worldwide, causing pain and stiffness in the joints. Arthroplasty is commonly performed to treat joint osteoarthritis, reducing pain and improving mobility. Nevertheless, a significant share of patients remain unsatisfied with their surgery. Personalised arthroplasty was introduced to improve surgical outcomes however current solutions require delays, making it difficult to integrate in clinical routine. We propose a fully automated workflow to design patient-specific implants for total knee arthroplasty.

\textbf{Methods} The proposed pipeline first uses artificial neural networks to segment the femur and tibia proximal and distal extremities. Then the full bones are reconstructed using augmented statistical shape models, combining shape and landmarks information. Finally, 77 morphological parameters are computed to design patient-specific implants. The developed workflow has been trained on 91 CT scans and evaluated on 41 CT scans, in terms of accuracy and execution time.

\textbf{Results} The workflow accuracy was $0.4\pm0.2mm$ for segmentation, $1.0\pm0.3mm$ for full bone reconstruction, and $2.2\pm1.5mm$ for anatomical landmarks determination. The custom implants fitted the patients' anatomy with $0.9\pm0.5mm$ accuracy. The whole process from segmentation to implants' design lasted about 15 minutes. 

\textbf{Conclusion} The proposed workflow performs a fast and reliable personalisation of knee implants, directly from a CT image without requiring any manual intervention. It allows the establishment of a patient-specific pre-operative planning in a very short time, making it easily available for all patients. Combined with efficient implant manufacturing techniques, this solution could help answer the growing number of arthroplasties while reducing complications and improving patients' satisfaction.


\section{Introduction}

Osteoarthritis (OA) affects about 528 million people worldwide, causing pain, swelling and stiffness in the joints \cite{WHO2023}. Total joint arthroplasty is commonly performed to treat OA, when the joint is too severely affected and palliative approaches can no longer alleviate pain or improve mobility. Joint replacement surgery has evolved significantly since its conception. It gained acceptance with ever-improving implant survivorship and now aims to provide a ‘forgotten joint’ to most patients. To do so, personalising joint replacement is the key solution to restore native joint kinematics, function, and perception \cite{Vendittoli2023}. Continuous research efforts are conducted in that sense, mostly for hip and knee replacements. Indeed, they are by far the most performed arthroplasties nowadays, followed by shoulder, elbow, foot/ankle and hand/wrist replacements \cite{Katano2021,NZOA2021}. \textcolor{black}{Although total knee arthroplasty (TKA) can provide excellent clinical results, it suffers from a high number of patients who remain dissatisfied with the results of the surgery compared with total hip arthroplasty\cite{Benignus2023}.}

\textcolor{black}{On a first hand, the main challenge of TKA in its early years, was to improve the implant stability. Consequently, surgical techniques such as the mechanical alignment (MA) were introduced to simplify and standardise the operations \cite{Riviere2020book}. MA also aimed to maximise the longevity of the TKA prosthesis by balancing load distribution between the implants' compartments \cite{Minoda2023}.}
\textcolor{black}{The development of assistive technological tools, such as navigation, robotics and patient-specific instrumentation, improved the accuracy of implantation with respect to the MA strategy, enhancing surgical outcomes as well as the satisfaction rate among TKA patients.}
Between 85\% and 95\% of them were satisfied with their surgery over the last decade, when the satisfaction rate varied between 75\% and 89\% in the 2000s \cite{Riviere2020book,DeFrance2023}. Then, between 5\% and 15\% of TKA patients are still unsatisfied with their surgery outcomes, mainly because of their increased expectations regarding symptoms, physical function, quality of life, coping strategies, and longevity of implant \cite{Lutzner2023}. Several causes can be identified as source of dissatisfaction: overhang of the femoral component causing pain (almost 27\% of all clinically important pain \cite{Mahoney2010}); abnormal biomechanics, related to the cruciate ligaments removal as well as to the lack of consideration of the patient anatomy, limiting the mobility and stability of the knee \cite{Schmidt2003}; or implants early mechanical fail, due for most to positioning, alignment or fixation defects \cite{Sharkey2014}. \textcolor{black}{In the end, such poor post-operative outcomes are likely to require revision surgery.}

\textcolor{black}{On a second hand, the number of TKA procedures has been rising steadily since the 1990s in Europe and the USA, and this increase is forecast to continue for the next decades \cite{LeStum2022,Shichman2023}. 
For instance, recent predictions for 2050 expect 150 000 TKAs in France (versus 102 655 in 2019), and more than 1 800 000 TKAs in the USA (versus 480 958 in 2019)\cite{LeStum2022,Shichman2023}.}
With a revision rate varying between 2\% and 12\% \cite{Erivan2020,Katano2021,SAR2021,AAOS2022}, the number of revision TKA is also expected to grow until 2050 and so even more rapidly than primary TKA \cite{Klug2021}. Two main factors can partially explain this growth: the ageing population and the expansion of TKA indication to younger and less severe symptoms patients \cite{LeStum2022}. While in 1997 patients younger than 65 years old represented 25\% of TKA, they now represent 40\% \cite{Pabinger2015}. These patients have longer life expectancy, and so have higher risks to require a revision. Moreover, TKA for low grade OA represents higher costs than TKA for severe OA \cite{Losina2015}. Therefore, TKA will inevitably represent a huge burden for the future health care systems \cite{Klug2021}. 

\textcolor{black}{To answer TKA increasing demand and reduce the need for revision, surgeries have to be made faster while being more reliable.}
Consequently, TKA’s main focus is now shifting from enhancing implant survivorship toward improving patient function, kinematics and satisfaction. Currently, alternative techniques to MA are gaining interest, \textcolor{black}{such as anatomical or kinematic alignment}, to better respect the wide range of normal anatomy of the knee and restore individual anatomy with a personalised joint replacement \cite{Nisar2020}.  
\textcolor{black}{These alignment techniques all aim at resurfacing the femorotibial joint, matching the implant geometry with the bony anatomy to restore the native pre-arthritic limb alignment and adjust the ligament balance \cite{Minoda2023,Lustig2021}}.

With TKA patients being more active today than ever, performing TKA that mimic the natural knee is therefore essential to the patients’ long-term satisfaction and survival \cite{Li2017}. To restore kinematic function, implant customisation aims to correct the patient’s knee deformities while staying as close as possible to his anatomy \cite{Lee2020,Bonnin2022}. Custom implants offer three features that are rarely attainable when using off-the-shelf (OTS) implants: (1) Optimisation of the implant-bone fit, to avoid overhang or under-coverage; (2) Decoupling of the patellofemoral and tibiofemoral compartment, to optimise patellofemoral and tibiofemoral kinematics independently; and (3) Restoration of native condylar curvature, to improve ligament balancing, mid-flexion stability and kinematics \cite{Saffarini2023,Sappey-Marinier2020}. Although there is still no consensus on the advantages of custom implants compared to OTS implants in terms of satisfaction and pain scores \cite{Victor2021,Wendelspiess2022,BeitNer2021}, custom TKA demonstrated significant benefits regarding  over- and under- sizing, local tendon impingement, improved kinematics, lower complication rate, and facilitation of restoration of constitutional coronal alignment \cite{Victor2021}. However, replicating the patient anatomy is not sufficient to ensure the implant stability and patient satisfaction, as the cruciate ligaments are removed during the surgery and can no longer stabilise the joint. Combining custom TKA with latest prosthetic concepts and ‘personalised alignment’ is expected to improve patient-reported outcome measures compared to OTS TKA \cite{Gousopoulos2023} and represents one of the current biggest challenges to perform personalised TKA \cite{Victor2021}.

Custom TKA implants have the potential to greatly improve knee kinematics and patient knee functions compared to OTS TKA implants. However, further investigation is needed to make the custom TKA implant readily accessible for patients \cite{Li2017}. \textcolor{black}{Today, only two solutions are proposed on the market of personalised implants:} Origin\textsuperscript{\textregistered} custom TKA (Symbios, Yverdon-les-Bains, Switzerland) and iTotal\textsuperscript{TM}CR G2 (ConforMIS Inc., Bedford, MA, US). Both solutions provide a personalised implant along with patient-specific instrumentation designed based on a computed tomography (CT) scan of the patient’s lower limb. The personalised implants and instrumentation are delivered in a ready-to-use box, thus reducing the stock of implants and instrumentation needed in situ. However, the implant customisation is based on manual and semi-automated methods, making the design and manufacturing process last for up to 8 weeks\textcolor{black}{, from planning to delivering the final implant} \cite{Sappey-Marinier2020,Bonnin2022,Gousopoulos2023}. \textcolor{black}{Furthermore, establishing the pre-operative planning and designing custom implants require several back and forth between surgeons and the engineering team, making it difficult to personalise surgery on a day-to-day basis.} Such processes need to be automated to allow a simple and fast establishment of the pre-operative planning for a seamless integration in clinical routine\textcolor{black}{\cite{Luo2024}} and make personalised TKA available for patients needing surgery within less than two months. \textcolor{black}{Recent research efforts have been undertaken to automatise implants personalisation, but are limited by one or more of the following factors: they have focused solely on the femoral component, did not consider knee alignment, have been tested on a single study case, or have involved semi automated or manual processes\cite{Ghidotti2024,Burge2023}}.

Therefore, to the best of our knowledge, we propose the first fully automated workflow to design patient-specific TKA implants from CT images.

\section{Materials and methods}

\subsection{Global workflow}

\begin{figure*}[ht]
    \centering
    \includegraphics[width=\linewidth]{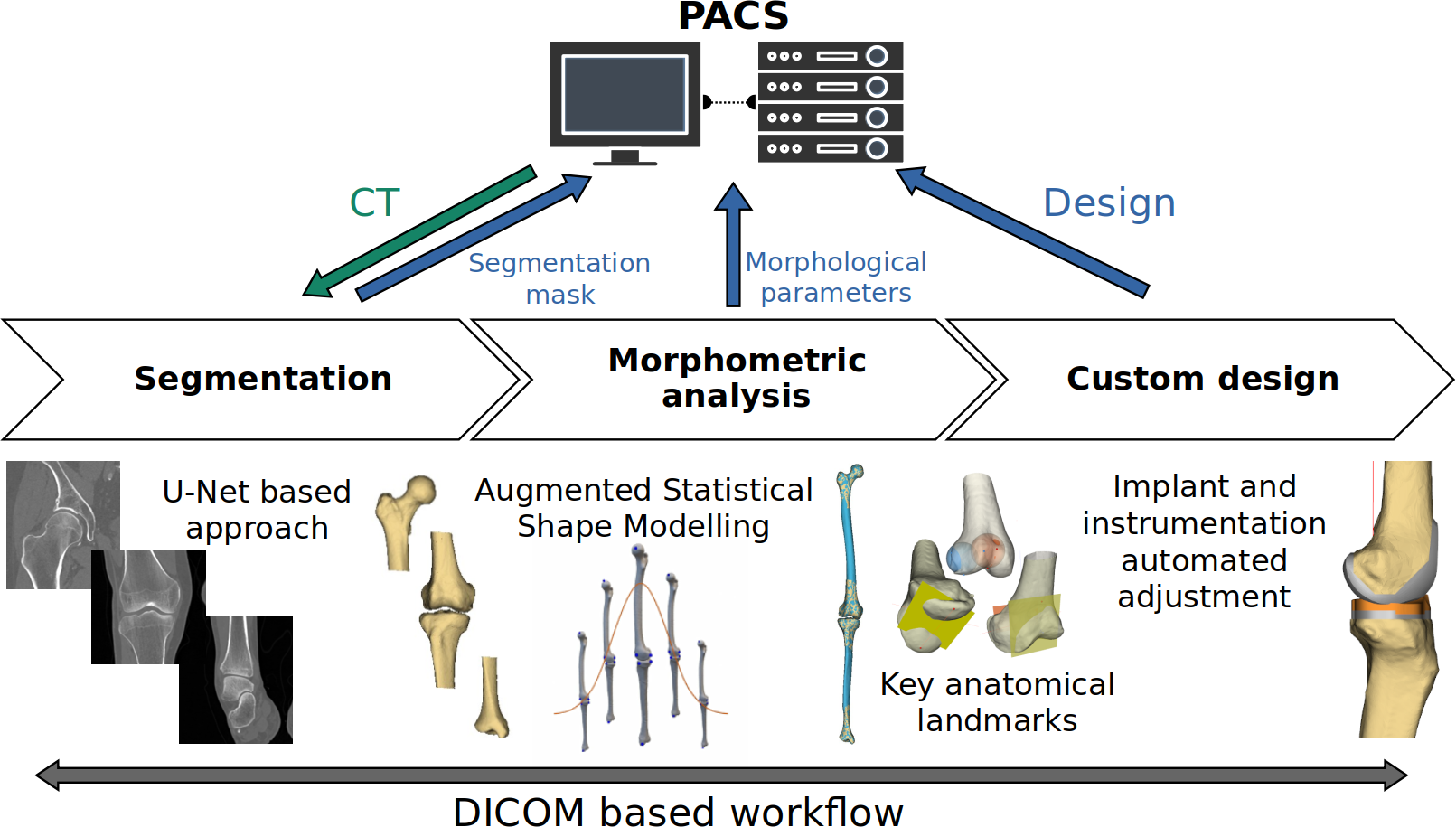}
    \caption{\textbf{Global workflow} overview to generate patient-specific orthopaedic implants.}
    \label{fig:global-workflow}
\end{figure*}

We developed a pre-operative planning software that can be connected to a PACS (Picture Archiving and Communication System) for direct access to the patients’ data. The different steps of the planning can be performed automatically within the software: segmentation of the hip, knee and ankle joints, determination of key anatomical landmarks and design of patient-specific implants. The whole workflow is easily traceable, as the result of each step is saved directly within the PACS under the patient ID and complies with DICOM format. An overview of the global workflow is proposed in Fig \ref{fig:global-workflow}.

\subsection{Dataset}

132 anonymised CT scans of lower limbs (right and/or left) from 75 patients were collected on 22/10/2021. These scans were acquired from 3 different machines (Toshiba\textsuperscript{\textregistered} Aquilon One, Siemens\textsuperscript{\textregistered} Somatom Edge Plus, Siemens\textsuperscript{\textregistered} Somatom Definition Edge) at the Brest University Hospital during the years 2020-2021. 
Approval for this study was obtained from the Medical Ethics Committee of the University Hospital of Brest that coordinated the project under the Kneemod trial (N°29BRC18.0235), and is registered in Clinical Trial (NCT04179812). Participants gave informed written consent.
\textcolor{black}{Patients were included if CT images of their leg were present in the University Hospital of Brest database, they had knee trauma, osteoarthritic or healthy knee, and they formulated non opposition.
Patients were excluded if they refused to participate or if only partial views were available, thus not allowing a good segmentation of the total joints.}
The volumes were manually segmented and then cropped to create three separate datasets for each joint: hip, knee and ankle. The data was split into training (91 scans from 50 patients) and test sets (41 scans from 25 patients).

\subsection{CT image segmentation}

A dedicated segmentation algorithm, integrating deep learning and image post processing principles, was developed to automatically extract the patient’s bones from CT images.

In order to achieve accurate and fast segmentation suitable for clinical usage in TKA planning, we developed three convolutional neural networks (CNNs) exclusively based on 2D data. Each model is dedicated to a specific joint (ankle, knee or hip), and takes a series of transverse cross-sections centred on the bone as input. All models were constructed following the principles of the  2D U-Net \cite{Ronneberger2015}. The global architecture, parameters and data formatting were similar to those described in \cite{Maintier2023}.

\textcolor{black}{The predicted 2D segmentation masks were post-processed with morphological filtering and a watershed algorithm \cite{Beucher2018} to ensure femur and tibia separation in ambiguous cases.
} Finally, the 2D segmentation masks were stacked to construct 3D meshes of the bones, using the marching cubes and smoothing algorithms \cite{Lorensen1998}.

The accuracy of the automated segmentation was evaluated by comparison to reference segmentation (performed manually) using different metrics: the Dice Coefficient (DC) on the segmentation masks, the root mean square error (RMSE) and the Hausdorff Distance (HD) on the reconstructed 3D meshes. Also, the segmentation processing time was recorded. 

\subsection{Morphometric analysis}

\subsubsection{Augmented SSM building and fitting}
The full shape of the patient’s femur and tibia were reconstructed by fitting statistical shape models (SSMs) to the automatically segmented distal and proximal extremities of the bones.

Two SSMs have been built, one for each bone, from the training dataset. All left femurs and tibias were initially flipped around their longitudinal axis to obtain only right bones datasets. Both femur and tibia SSMs have been built following the workflow previously described in \cite{GuezouPhilippe2022mbec}. First, a virtual reference shape is computed to establish unbiased correspondence between the training data. \textcolor{black}{Then, a principal component analysis is performed on the training data in correspondence to compute the SSM's mean shape and modes of deformation.} 
The SSMs were augmented by integrating information on anatomical landmarks to the model (see next section for more details).

\textcolor{black}{To reconstruct the whole femur and tibia bones from their proximal and distal extremities, a custom algorithm was developed to fit a SSM to partial data. Firstly, the SSM mean shape is scaled to match the partial data bounding box's lengths. Secondly, the scaled SSM mean shape is rigidly registered to the partial data. Thirdly, the SSM is deformed to fit the bone surface as much as possible without reproducing osteophytes.
}
\subsubsection{Extraction of morphometric data}

To further design the custom implant, several anatomical landmarks were automatically identified.
The anatomical landmarks of the deformed SSM (inherited from the landmarks identified on the SSM mean shape) were used as initialisation for the landmarks detection. The landmarks were then adjusted to the segmented mesh to improve the detection accuracy. This adjustment was weighted regarding the risk of having osteophytes in the concerned zone.

A total of 77 morphological parameters (48 femoral and 29 tibial) were determined \textcolor{black}{on the bone 3D models}. 23 primary anatomical landmarks were directly detected with the fitted SSMs (16 femoral and 7 tibial landmarks), and 54 secondary parameters (anatomical landmarks, axes, planes, pointset, lengths or angles) were computed from these primary landmarks. 

\subsubsection{Validation}

The quality of both tibia and femur SSMs were evaluated by computing three metrics, as described by Davies et al. \cite{Davies2008}: compactness (ability to represent the variability of the training dataset), generality (ability to fit the testing dataset) and specificity (ability to generate shapes similar to the training dataset).

The full bone reconstruction accuracy was assessed by fitting the SSMs to the proximal and distal extremities of the 41 tibias and femurs of the testing dataset. The fitting duration has been registered and its accuracy has been evaluated by comparing the fitted models to the full bones manually segmented in terms of RMSE and HD.

\begin{figure*}
    \centering
    \includegraphics[width=\linewidth]{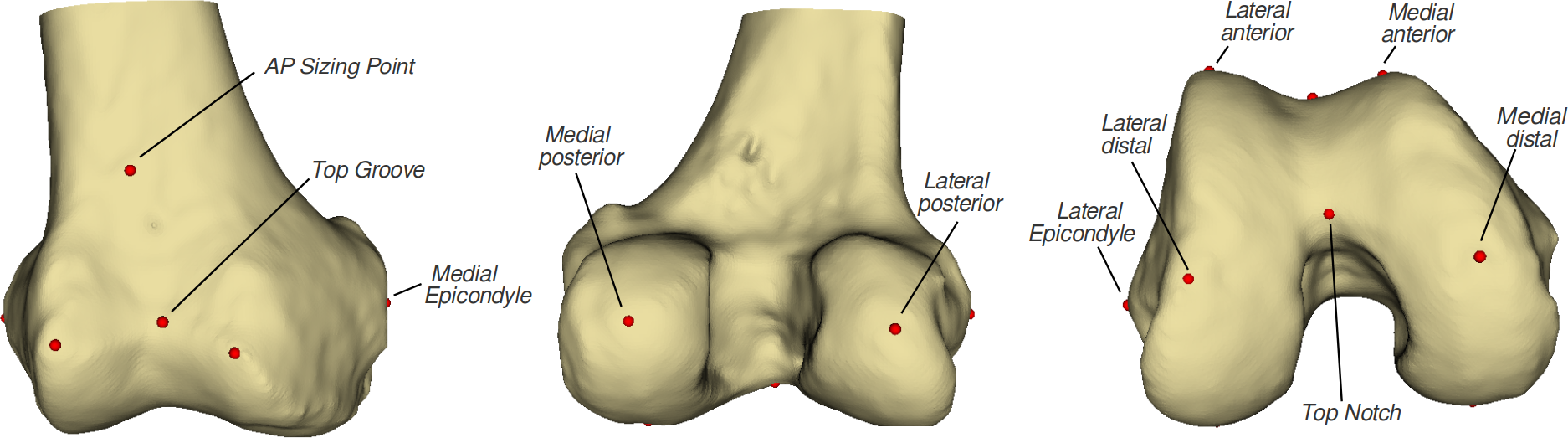}
    \\ \vspace{0.5\baselineskip} 
    \includegraphics[width=0.6\linewidth]{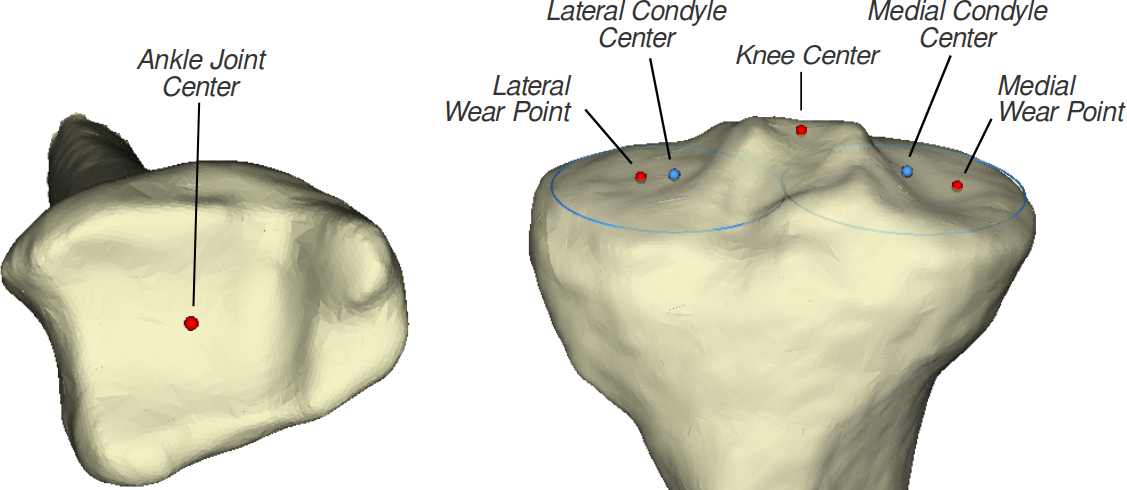}
    \caption{\textbf{Femoral and tibial landmarks acquired for the morphometric analysis validation.} The condyles centers were determined from 5 points acquired on the edge of the lateral and medial plateau (blue lines). }
    \label{fig:landmarks}
\end{figure*}

The landmarks computation was validated based on the analysis of 17 primary landmarks (11 femoral and 6 tibial) detailed in Fig \ref{fig:landmarks}. The computed landmarks were compared to ground truth landmarks acquired manually on the 41 shapes of the testing dataset. The 11 femoral landmarks were determined twice by two orthopaedic surgeons on 3 modalities : CT images, virtual 3D models, and 3D printed models. 
A previous study showed that no modality was more reliable than another and that the intra observer variability did not impact the computation of secondary parameters \cite{Clave2023}. The 7 tibial landmarks were thus determined twice by two orthopaedic surgeons on virtual 3D models only.
The ground truth for each landmark was defined as the barycenter of the corresponding landmarks picked manually. The computed landmarks accuracy was evaluated in terms of distance (in millimetres) to the ground truth landmarks.
As an exception, the tibial condyles centres were computed from 5 points picked on the edge of both the lateral and medial plateau, and their accuracy was evaluated through the angular error (in degrees) of the line they form.

\subsection{Implant design}
The entire design process is performed automatically using a sequence of custom operations in the open source software Open CASCADE Technology (Open Cascade, part of Capgemini, Issy-Les-Moulineaux, France). The implants are designed based on the patient’s anatomical landmarks previously determined and follow the medial pivot concept. Such design has been elected to restore the natural mobility of the knee when cruciate ligaments have been removed, with the knee rotating around the medial condyle and allowing antero-posterior translation of the lateral condyle \cite{Sabatini2018}.

\subsubsection{Femoral implant}
The design of the femoral implant starts with the posterior part and in first place the medial pivot sphere (Fig \ref{fig:implant-design}a). For each condyle, the radius of the medial pivot sphere is computed from the antero-posterior size \cite{Mahfouz2007}. 
The posterior surfaces of both medial and lateral condyles are initially shaped based on this sphere, after which they are refined to tightly align with the contours of the resected bone. The posterior most proximal part is then cut along a circle arc to obtain the roll back shape that will allow deep flexion. Secondly, the anterior surface is designed to closely adhere to the patient's premorbid anatomy, aiming to facilitate patellar tracking. The specific characteristics of each patient's lateral and medial facet height and sulcus angle are individually taken into consideration to prevent the risk of patellar instability \cite{Dejour1994,Chen2023}.

\subsubsection{Tibial implant}
\begin{figure*}
    a) \includegraphics[width=0.95\linewidth]{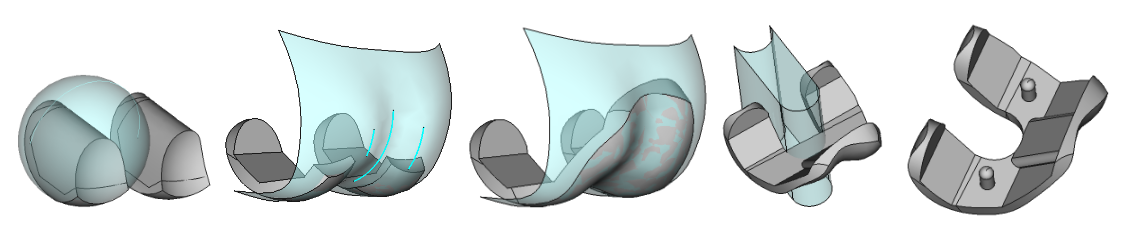} \\
    b) \includegraphics[width=0.8\linewidth]{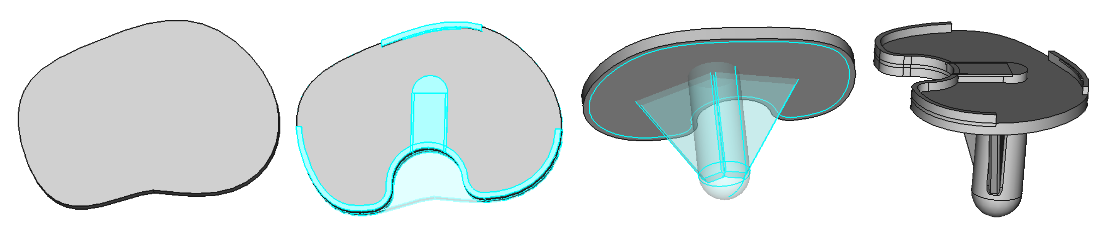}
    \caption{\textbf{Design steps for the personalised prosthesis.} a) Femoral implant, from left to right: defining the medial pivot sphere; defining the condyles’ curvature; defining the anterior surface; cutting the intercondylar notch; adding pins and chamfers. b) Tibial implant, from left to right: tibial baseplate contouring; adding locking mechanism for the polyethylene insert; adding the keel; adding chamfers.}
    \label{fig:implant-design}
\end{figure*}

The shape of the tibial baseplate is crafted to conform to the contours of the resected bone, ensuring optimal contact with the cortical bone to enhance implant stability and durability (Fig \ref{fig:implant-design}b). The location of the keel is determined by taking into account the morphology of the diaphysis and the tibial plateau. The length of the stem, as well as the dimensions of the two vanes, are also functions of the patient’s specific characteristics.

\subsubsection{Tibial insert}
The polyethylene tibial insert is designed regarding the femoral and tibial implants. The basic shape of the insert is built from the embase shape of the tibial component. The footprint of the medial pivot sphere is first excavated into the basic shape. Then, the lateral part is designed almost flat, to allow free antero-posterior translation of the femoral component as suggested by the medial-pivot concept. Finally the most anterior part is cut to avoid conflict with the patella in deep flexion.

\subsubsection{Design evaluation}
\begin{figure}[h]
    \centering
    \includegraphics[width=0.3\linewidth]{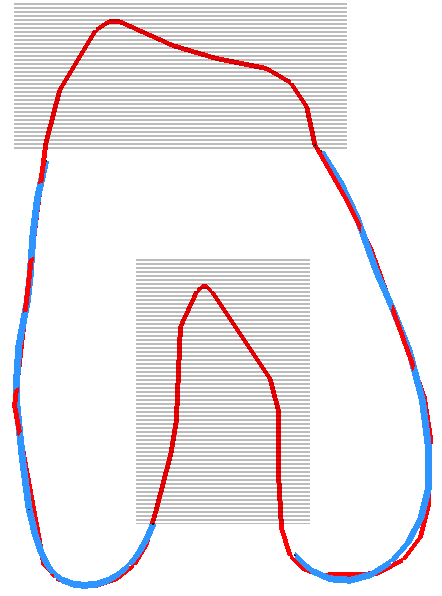}
    \includegraphics[width=0.4\linewidth]{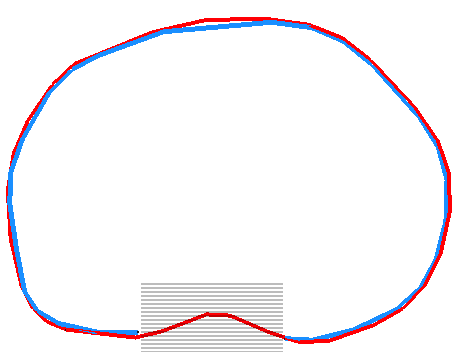}
    \caption{\textbf{Flattened contours used for over and under-hang computation.} The red and blue lines represent bone resection and implant contours respectively. Grey zones are ignored in the computation.}
    \label{fig:OUH}
\end{figure}

To evaluate the quality of the designed custom implants, the accuracy of the bone-implant fitting is computed by measuring over and under-hang, namely, the distance between the contours of the bone and the implant. Several zones were ignored: (1) the posterior part of the tibial implant, as the implant shape voluntarily does not follow the resection one; (2) the femoral intercondylar notch, as it is not relevant in the residual pain outcome of the surgery; and (3) the femoral anterior proximal part, as the implant does not try to reproduce the anatomy in this zone (see Fig \ref{fig:OUH}).

\section{Results}

\subsection{Segmentation of partial CT image}

One case has been excluded from the testing dataset because of too much flexion in the lower limb leading to segmentation failure. The following results are reported for the remaining 40 cases.

The ankle, knee and hip CT image were automatically segmented with a mean DC of $98.5 \pm 0.8\%$, mean RMSE of $0.4 \pm 0.3mm$ and mean HD of $3.0 \pm 2.1mm$. The segmentation accuracy for each bone is detailed in Table \ref{tab:resu-seg}. The computation time to fully segment all three joints of a patient's leg was $378 \pm 221$ seconds on a CPU (Intel\textsuperscript{\textregistered} Core\textsuperscript{\textsc{tm}} i7-7820HQ, 2.90GHz), depending on the volume depth.

\begin{table}[h]
\begin{adjustwidth}{-2.25in}{0in} 
    \centering
    \caption{\textbf{Segmentation accuracy.}}
    \label{tab:resu-seg}
    \begin{tabular}{llccc} \hline
Bone  & Extremity & DC (\%) & RMSE (mm) & HD (mm) \\ \hline
\multirow{2}*{Femur} & Proximal  & $98.6 \pm 0.9$ $[93.7 - 99.3]$ & $0.4 \pm 0.2$ $[0.3 -  1.8]$ & $3.0 \pm 1.7$ $[1.3 - 10.9]$ \\
                     & Distal    & $98.6 \pm 0.9$ $[95.5 - 99.4]$ & $0.4 \pm 0.4$ $[0.2 -  2.4]$ & $3.1 \pm 2.8$ $[0.9 - 15.1]$ \\ \grayline
\multirow{2}*{Tibia} & Proximal  & $98.7 \pm 0.5$ $[96.8 - 99.4]$ & $0.3 \pm 0.1$ $[0.0 -  0.5]$ & $2.8 \pm 1.3$ $[0.0 -  5.8]$ \\
                     & Distal    & $98.2 \pm 0.9$ $[94.2 - 99.1]$ & $0.4 \pm 0.3$ $[0.0 -  1.7]$ & $3.0 \pm 2.4$ $[0.0 -  9.5]$ \\ \hline
    \end{tabular} \\\vspace{2pt}
    \small{Dice coefficient (DC), root mean square error (RMSE) and Hausdorff distance (HD) are reported as mean $\pm$ std.dev. [min - max].}
\end{adjustwidth}
\end{table}

\subsection{Morphometric analysis}

\subsubsection{SSM quality}

The evaluation metrics of the femur and tibia SSMs are detailed in Table \ref{tab:resu-ssm}, with respect to the number of principal components (modes) kept by the model. The SSMs are later reduced to their first N modes of deformation representing at least 99\% of variability to simplify and speed up the SSMs fitting (i.e. 15 and 11 modes for the femur and tibia SSMs respectively).

\begin{table}[h]
    \centering
    \caption{\textbf{Tibia and femur SSMs metrics, regarding the number of modes (N).}}
    \label{tab:resu-ssm}
    \begin{tabular}{lcccc} \hline
        Bone & N & Compactness (\%) & Generality (mm) & Specificity (mm) \\ \hline
        \multirow{4}*{Tibia}&1 & 88.7  & $2.11 \pm 0.83$ & $1.39 \pm 0.21$ \\
        &3 & 95.9  & $1.72 \pm 0.82$ & $1.42 \pm 0.32$ \\
        &11 & 99.0 & $1.37 \pm 0.77$ & $1.84 \pm 0.70$ \\ 
        &89 & 100  & $1.26 \pm 0.39$ & $2.90 \pm 0.90$ \\ \hline
        \multirow{4}*{Femur} &1  & 84.4 & $2.30 \pm 0.78$ & $1.74 \pm 0.26$ \\
        &5  & 95.8 & $1.76 \pm 0.54$ & $2.39 \pm 0.60$ \\
        &15 & 99.1 & $1.41 \pm 0.68$ & $2.55 \pm 0.49$ \\ 
        &89 & 100  & $1.31 \pm 0.79$ & $1.84 \pm 0.41$ \\ \hline
    \end{tabular}\\ \vspace{2pt}
    \small{Generality and specificity are reported as mean $\pm$ std.dev.}
\end{table}

\subsubsection{Bone reconstruction}

The full femur and tibia bones were automatically reconstructed from partial data with a mean RMSE of $1.0 \pm 0.3mm$ and mean HD of $4.1 \pm 1.3mm$. The registration accuracy for each bone is detailed in Table \ref{tab:resu-reg}. The reconstruction of the femur and tibia lasted $329 \pm 55$ seconds.

\begin{table}[h]
    \centering
    \caption{\textbf{Full bone reconstruction accuracy.}}
    \begin{tabular}{llcc} \hline
Bone  &  Extremity  & RMSE (mm)  & HD (mm)  \\ \hline
\multirow{2}*{Femur} & Proximal  & $1.2 \pm 0.2$ $[0.9 - 1.5]$ & $4.5 \pm 0.9$ $[3.2 - 6.6]$  \\ 
                     & Distal    & $1.0 \pm 0.3$ $[0.7 - 2.1]$ & $4.1 \pm 1.3$ $[2.7 - 8.8]$  \\ \hline
\multirow{2}*{Tibia} & Proximal  & $1.0 \pm 0.3$ $[0.7 - 2.3]$ & $4.1 \pm 1.2$ $[2.8 - 9.1]$  \\ 
                     & Distal    & $0.8 \pm 0.3$ $[0.5 - 2.1]$ & $3.6 \pm 1.7$ $[1.9 - 10.5]$ \\ \hline
    \end{tabular}
    \label{tab:resu-reg}\\ \vspace{2pt}
    \small{RMSE and HD are reported as mean $\pm$ std.dev. [min - max].}
\end{table}

\subsubsection{Anatomical landmarks determination}

The anatomical landmarks have been determined with a mean accuracy of $2.2 \pm 1.5mm$ ($2.5 \pm 1.6mm$ and $1.5 \pm 1.2mm$ for the femoral and tibial landmarks respectively). 
The accuracy per landmark is detailed in Table \ref{tab:resu-landmarks}.

\begin{table}[h]
\begin{adjustwidth}{-2.25in}{0in} 
    \centering
    \caption{\textbf{Accuracy of the morphological parameters automatic determination.}}
    \label{tab:resu-landmarks}
\begin{minipage}{.45\linewidth}
\begin{tabular}{lcc} \hline
Femoral Landmarks           & mean $\pm$ std& [min - max]   \\ \hline
AP Sizing Point             & $3.0 \pm 1.7$ & $[0.3 - 8.2]$ \\
Lateral Anterior            & $1.9 \pm 1.2$ & $[0.2 - 5.3]$ \\
Lateral Distal              & $4.0 \pm 1.8$ & $[0.8 - 7.9]$ \\
Lateral Epicondyle          & $2.2 \pm 1.2$ & $[0.5 - 6.2]$ \\
Lateral Posterior           & $2.8 \pm 1.4$ & $[0.3 - 6.3]$ \\
Medial Anterior             & $1.3 \pm 0.8$ & $[0.3 - 4.0]$ \\
Medial Distal               & $3.1 \pm 1.8$ & $[0.2 - 9.3]$ \\
Medial Epicondyle           & $3.0 \pm 1.8$ & $[0.2 - 7.6]$ \\
Medial Posterior            & $1.8 \pm 0.8$ & $[0.4 - 3.4]$ \\
Top Groove                  & $2.2 \pm 1.2$ & $[0.5 - 4.8]$ \\
Top Notch                   & $1.9 \pm 1.0$ & $[0.4 - 4.7]$ \\ \hline
\end{tabular}
\end{minipage}
\begin{minipage}{.45\linewidth}
\begin{tabular}{lcc} \hline
Tibial Landmarks            & mean $\pm$ std& [min - max]   \\ \hline
Ankle Joint Center          & $2.5 \pm 1.0$ & $[0.9 - 5.1]$ \\
Knee Center                 & $2.2 \pm 0.9$ & $[0.7 - 4.1]$ \\
Lateral Wear Point          & $0.7 \pm 0.6$ & $[0.0 - 2.5]$ \\
Medial Wear Point           & $0.7 \pm 0.6$ & $[0.0 - 2.5]$ \\ \grayline
Condyle Centers Joint Line  & $2.0 \pm 1.6$ & $[0.0 - 6.2]$ \\ \hline
\end{tabular}\\ \vspace{3.8\baselineskip} \\
\small{Errors in mm, except for the tibial condyle centers line angular error in $^\circ$.}
\end{minipage}
\end{adjustwidth}
\end{table}

\subsection{Implant design}
Designing the custom implants succeeded in 89\% of cases and failed in 4 cases (2 femoral implants, 1 tibial implant and 1 PE insert were invalid). 
The computation of the custom design for both femoral and tibial implants lasted $270 \pm 80$ seconds. The implants fitted the bones with an RMSE of $0.9 \pm 0.5mm$ and HD of $3.1 \pm 2.1mm$ (over and under hang). More details on the implant-to-bone fitting can be found in Table \ref{tab:resu-ouh}. The highest hausdorff values corresponded to under hang at the posterior lateral tibial plateau. Such coverage was obtained due to a retreat of the plateau in the design to avoid muscle impingement.

\begin{figure}[h]
    \centering
    \includegraphics[width=0.2\linewidth]{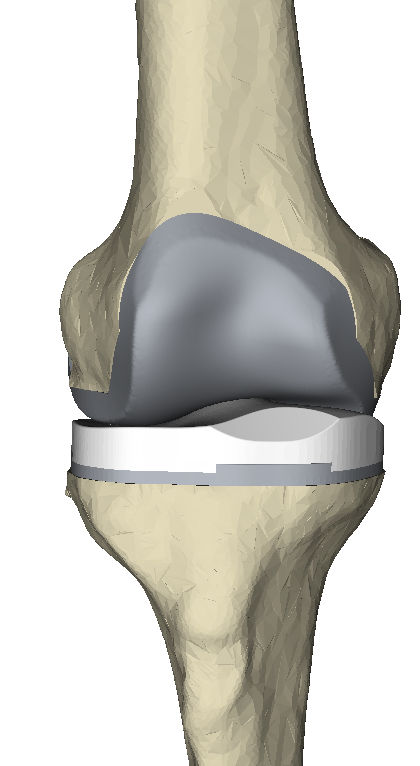} \quad
    \includegraphics[width=0.2\linewidth]{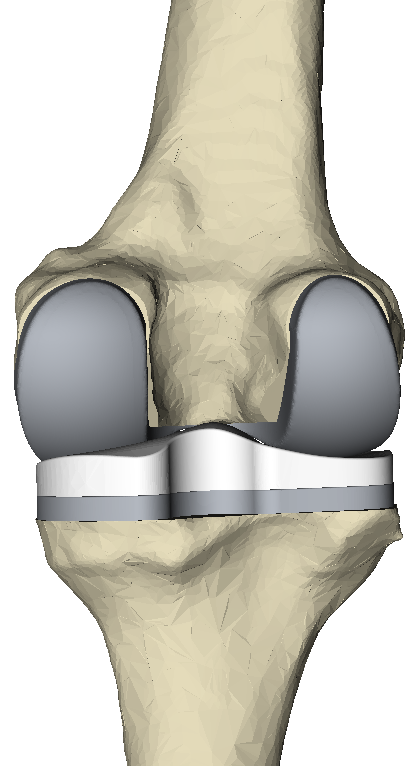} \quad
    \includegraphics[width=0.2\linewidth]{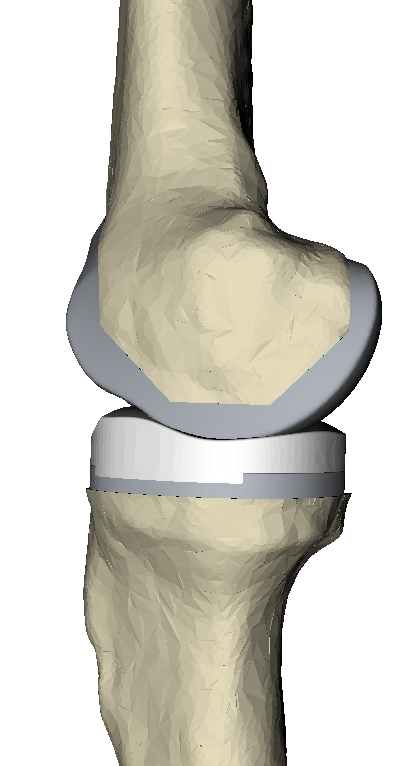} \\
    Anterior view \qquad Posterior view \qquad Sagittal view \\
    \includegraphics[width=0.2\linewidth]{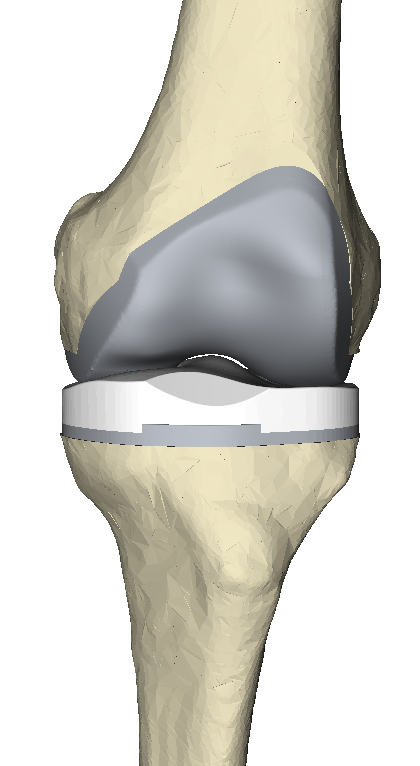} \quad
    \includegraphics[width=0.2\linewidth]{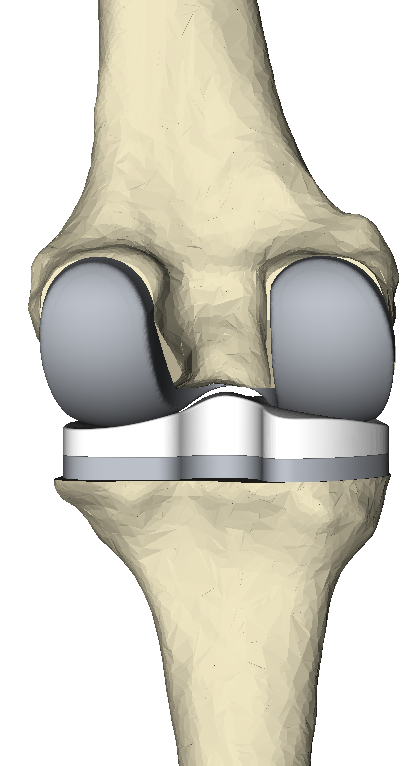} \quad
    \includegraphics[width=0.2\linewidth]{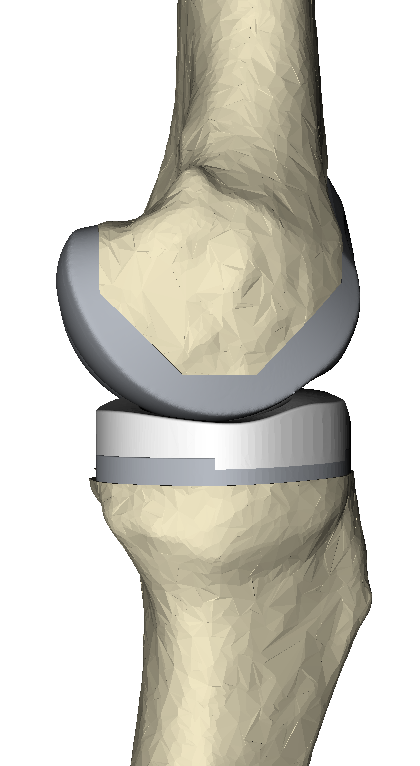}
    \caption{\textbf{Example of personalised implants designed for two patients} for their right (top row) and left (bottom row) knees.}
    \label{fig:ex-implants}
\end{figure}

The custom implants for two patients are illustrated in Fig \ref{fig:ex-implants}.

\begin{table}[t]
    \centering
    \caption{\textbf{Bone coverage for custom design implants (over and under hang).}}
    \label{tab:resu-ouh}
    \begin{tabular}{lcc} \hline
Bone  & RMSE (mm)                   & HD (mm)                     \\ \hline
Femur & $0.4 \pm 0.1$ $[0.3 - 0.6]$ & $1.2 \pm 0.4$ $[0.7 - 2.8]$ \\
Tibia & $1.4 \pm 0.3$ $[0.9 - 1.9]$ & $4.9 \pm 1.3$ $[2.5 - 7.1]$ \\ \hline
    \end{tabular}
    \\ \vspace{2pt} \small{RMSE and HD are expressed as mean $\pm$ std.dev. [min - max].}
\end{table}

\section{Discussion}

The proposed method automatically designs patient-specific TKA implants in about 15 minutes, mimicking the joint anatomy and respecting the limb kinematics, by using the medial pivot prosthetic concept, to ensure the proper motion and stability of the knee.

\paragraph{Segmentation accuracy.}
The mean segmentation accuracy was $0.4 \pm 0.3mm$. We observed high HD on the ankle often resulting from inaccurate detection of the medial malleolus extremity. However, the absence of this portion of the tibial bone does not significantly affect the subsequent planning process. Indeed, the ankle's primary role is to help determine the mechanical axis of the tibia, which runs from the knee center located on the proximal tibia to the distal tibial plafond articular surface. This axis orientation remains relatively consistent even if the SSM fits a slightly shorter length for the ankle portion.

Likewise, we observed high HD for segmentation accuracy in the hip region when the great trochanter's extremity was inaccurately detected. However, these inaccuracies does not significantly impact the positioning of the mechanical axis of the femur after fitting the SSM. (The femoral mechanical axis runs from the center of the femoral head to the knee center located on the distal femur.)

The segmentation accuracy of our method is in line with similar studies in the literature. For instance, other U-Net derived methods, for knee segmentation on magnetic resonance images (MRI) or on CT images, reach a mean DC around 98\% with either bigger or smaller training and testing dataset \cite{Ridhma2021,Rossi2022,Kuiper2022}.

Lately, a lot of research has been conducted to develop the best architecture of deep-learning models to obtain segmentations as accurate as possible \cite{Wang2022}. Most of these algorithms yield excellent results, with small differences in Dice scores. However, for the major part, these models have been trained and tested on selected images, from non-pathological patients, with reasonably good image resolution and without interfering elements such as metal implants. Those selected images ensure an improved segmentation of the patient anatomy, although they are not representative of the clinical reality. Indeed, patients undergoing TKA usually suffer from high grade knee osteoarthritis, making the joint segmentation harder because of the highly damaged cartilage and presence of osteophytes. Current challenges for automated segmentation now include performing accurate segmentations whatever the patient’s morbidity, the presence of foreign bodies and image resolution.

\paragraph{Morphometric analysis.}
The proposed method automatically determines anatomical landmarks with a mean accuracy of $2.2 \pm 1.5mm$. Such errors are similar to the intra and inter operator variability existing when manually picking the landmarks \cite{Clave2023}, indicating that the proposed workflow is as accurate as manual acquisitions. The highest errors were obtained for the lateral distal point (most distal point on the femoral lateral condyle). This landmark was found to be the hardest to identify as it is situated in a flat area. In the proposed workflow, the lateral distal point is used to determine the femoral distal resection plane only. However, the flat area around the lateral distal point is parallel to the this resection plane. Therefore, errors in this point detection have negligible impact on the resection plane orientation and the resulting implant design.

The obtained results are in line with previous studies using similar or different approaches. 
Fischer et al. \cite{Fischer2020} proposed to automatically determine femoral landmarks, axes, planes and bone coordinate systems by morphing a femur template to a subject’s femur. They evaluated their method on 22 femoral landmarks based on manual acquisitions performed 4 times by 5 experts. They obtained excellent intra and inter observer reliability (min ICC=0.933), with manual median errors ranging from $0.3mm$ to $4.4mm$ depending on the considered landmark. Their automated method determined femoral landmarks with median errors ranging from $0.4mm$ to $6.7mm$. 
In the same way, Kuiper et al. \cite{Kuiper2023} fitted a mean bone model to 20 non pathological subjects to determine 32 femoral and tibial landmarks. They reached a mean accuracy of $2.17 \pm 1.37mm$ similar to their intra observer error ($2.01 \pm 1.64mm$).
Similarly, Chen et al. \cite{Chen2022} fitted a SSM to 141 non pathological male femora to determine the position of 6 anatomical landmarks. Their automatic method had an average error of $1.3mm$ similar to their intra and inter observer errors ($1.25mm$ and $1.29mm$ respectively). They obtained better accuracy than our method, however they tested their method on their training dataset, which eases the fitting of the SSM and thus the landmarks’ determination.
Deep-learning approaches have also been used to automatically determine bony landmarks. Yang et al. proposed a convolutional neural network to determine 7 landmarks of the distal femur and obtained mean errors around $5mm$ \cite{Yang2015}. More recently, Wang et al. developed a network based on PointNet++ and obtained errors below $5mm$, except in severe knee joint wear patients where the feature points become challenging to extract \cite{Wang2023}. 

Whatever the method used to detect the bony landmarks, the method’s validation and the definition of ground truth landmarks can raise certain concerns. Indeed, automatic methods are compared to manual measurements for validation. However, there exists intra and inter operator variability in determining anatomical landmarks, and with such variability, the ground truth landmarks can only be approximated. In the literature as in our study, reference landmarks are commonly defined as the median or mean of manual acquisitions to approximate the ground truths. As no information is available on the actual ground truth, the choice of these reference points can impact the method’s validation results. Nevertheless, as the errors introduced by the intra and inter operator variabilities are similar to random noise, averaging the manual acquisitions reduces the fluctuations around the ground truth landmarks, allowing a more accurate approximation. 

\paragraph{Implants custom design.}
Compared to off-the-shelf implants, personalised TKA implants improve the components fit, the kinematic function and consequently patient satisfaction, reduce medical complications and thus lower the procedure associated costs \cite{Lee2020,Beckmann2021}.

We chose to design our implant following the medial pivot concept which models the femoral condyles with two spheres, thus approximating the sagittal curvature with only one radius for each condyle. The implants design plays an essential role in joint kinematics, and consequently in restoring stable and natural knee movements. For instance, the femoral condyles curvature is a key factor as it impacts the tibiofemoral internal-external rotation and antero-posterior translation \cite{Asseln2021}. In this regard the J-curve design has been introduced to efficiently reproduce the femoral condyles sagittal curvature, with different radii of curvature. 
While the J-curve approach is by essence more anatomical than the medial pivot one, the latter has shown a greater ability to reproduce the natural knee biomechanics \cite{Kour2023}.
Even though the medial pivot design is still rarely used in practice \cite{Wittig2023} it is gaining more and more interest, with an increasing number of research articles being published.
Recent reviews show that medial pivot TKA is a reliable long-term treatment option for individuals with end-stage osteoarthritis, offering exceptional survivorship, low complication rates, and notable enhancements in clinical and functional outcomes \cite{Alessio-Mazzola2022,Cacciola2023}.

Designing personalised implants based on the patient's anatomy also raises multiple questions on the anatomy to be restored. When the objective is to recreate the patient's healthy joint, the implants can be designed based on the pre-morbid anatomy. However, still today, it is difficult to know where to place the threshold between pre-morbid and pathological anatomy and to what extent deformities should be reproduced or corrected remains unclear. 
For instance, most of the commercialised implants currently available recreate healthy femoral condyles but have flattened trochlea with high sulcus angle which characterises trochlear dysplasia \cite{Dejour2014}.
Moreover, there is no clear consensus on the advantage of reproducing the pre-morbid knee alignment using the kinematic alignment and its different versions (inverse, restricted, modified) in terms of clinical scores \cite{Lustig2021,Minoda2023}.
Now that the current technologies make it easier to design patient-specific implants, prospective clinical studies are needed to assess the short and long term outcomes of personalised TKA. Such research could help in determining the best prosthetic joint anatomy to offer a stable and functional knee, and to what extent should the deformity be corrected.

\paragraph{Global workflow.}
To our knowledge, only one article in the literature also proposed a complete automatic workflow to generate personalised implants for TKA from CT images. 
Burge et al. \cite{Burge2023} recently proposed a similar pipeline to ours, using machine learning to segment both the femur and the tibia, and statistical shape modelling to reconstruct the bones 3D models. They obtained custom implants models in less than 10 minutes, by restraining the SSM fitting to very smooth deformations and without determining any anatomical landmark. However, the custom implants they proposed may not be viable in the clinics. Indeed, their implants are directly carved out of the bones 3D models, the implant outer layer corresponding to the surface of the trimmed bone. Such implants replicate the anatomy of the bones -- whatever the deformities present -- but do not consider any kinematics aspect of the joint, such as the lower limb alignment, the knee flexion axis or the condyles congruence.
The personalised implants we propose, are designed regarding more than 70 parameters describing the lower limb anatomy and kinematics, and hence, are better suited for a clinical usage.

\textcolor{black}{Semi automated procedures have been proposed in the literature to shorten segmentation, pre-operative planning or implant design \cite{Ghidotti2024,Lambrechts2022,Stanca2023}, but the required manual interventions remain tedious and time-consuming which limits their usability in clinical routine.
Today, customisation solutions available on the market are still very little used. For instance, 10 000 custom implants from Symbios Orthopédie SA were implanted between 2018 and 2022, which represent less than 3\% of the TKAs in France on the same period (more than 300 000)\cite{LeStum2022,Symbios}.
Consequently, automating and simplifying all the processes from segmentation to implant manufacturing is needed to perform personalised TKA routinely.
The proposed approach for automating CT images segmentation, morphological analysis, and implant design, not only reduces the time needed for preparing custom TKA, but also enables surgeons to control the entire planning process, from start to finish, and to fine-tune it within a minute if necessary. Such automation simplifies the planning customisation, as it reduces the need to go back and forth with the implant designers and manufacturers to test different configurations, modifying one parameter at a time.}

\paragraph{Limitations.}
It is important to note that the dataset used in our study contains limited instances of severe osteoarthritis, which may not be representative of all patients undergoing TKA. \textcolor{black}{The inclusion criteria were designed to be not too restrictive in order to recruit a large variety of patients. However, it was collected from only one center which may be representative of a local population rather than the global population.
Therefore, the training and testing dataset should be increased to include CT images of more diverse patients to better establish the workflow applicability to a broader population in the context of surgical planning.}

A total of 77 morphological parameters are automatically determined on the patient’s femur and tibia. The accuracy of this morphological analysis was conducted on 17 femoral and tibial primary landmarks only. The remaining primary landmarks, such as the femoral head center, were not considered here as they cannot be picked manually on 3D printed or virtual meshes. The accuracy of secondary parameters was not analysed here, as secondary parameters are derived from primary ones and are directly link to the primary landmarks accuracy. Moreover, secondary parameters' ground truth may vary according to the surgeon preferences and the surgical plan adopted (e.g. resection plans orientation and depth).

The proposed workflow does not include the production of the custom designed implants. Such step could be performed in a time-effective manner using 3D printing techniques \cite{Meng2023}. However, whatever the technique used, manufacturing the implants usually involves a third-party supplier, which may add delays between the pre-operative planning and the surgery. Nevertheless, designing custom implants manually remains a time-consuming task that could be swiftly addressed with the proposed workflow. 

Such personalised prostheses require a higher accuracy than off-the-shelf prostheses when positioning the implants. Indeed, the implants are designed together with the pre-operative planning, ensuring the perfect fit of the implants to the operated bones. During preliminary experiments on cadavers, we observed that inaccuracies on the cutting planes' positioning would affect the quality of implant-to-bone fitting. Patient-specific cutting guides as well as computer assisted solutions (robotic arm or navigated instruments) could help improve the surgical accuracy and ensure the right positioning of the personalised implants.  

Finally, a mechanical evaluation is missing to ensure the stability and survivorship of the personalised implants and to validate the design of the polyethylene tibial insert. Work is currently ongoing to evaluate the implants resistance to fatigue and meet the regulatory standards for future commercialisation.

\paragraph{Perspectives.}
Experiments on cadavers have been performed as a first step to evaluate the complete workflow from designing the implants to performing the implantation. Future clinical studies are planned to assess the clinical benefit of the patient-specific implants designed automatically.

The proposed workflow can be easily adapted to other joint arthroplasties, once the relevant anatomical landmarks are identified. 

\textcolor{black}{Finally, a cost analysis will be carried out in the coming years to validate the feasibility of the proposed solution and subsequently integrate it into clinical practice. To this day, current solutions for producing custom implants require engineering work to process the CT images, model the patient's anatomy and design the implants. Such work represent a significant cost in the manufacture of implants, which may vary from a company to the other, and a country to the other.
The proposed solution automates these processing, modelling and designing steps, considerably reducing the time and engineering work required to carry them out. As a result, the associated costs will also fall significantly.}

\section{Conclusion}
We propose a complete and automated workflow to design custom TKA implants in a time effective manner. 
A dedicated U-net algorithm and SSMs were developed to automatically segment the patient's hip, knee and ankle and reconstruct the femur and tibia anatomy. A custom pipeline was implemented to determine more than 70 anatomical parameters and further design the patient-specific implants.
Such an approach is a key factor to increase the accessibility to personalised TKA, and more generally to personalised orthopaedic surgery to improve the surgical outcomes for all patients.

\section*{Declaration of competing interest}
This work benefited funding from the French government via the national research agency as part of the Investments for the Future Programs, under the reference ANR-17-RHUS-0005 (FollowKnee Project). The funders had no role in study design, data collection and analysis, decision to publish, or preparation of the manuscript.
No commercial funding was received for this study.
All authors declare that they have no known competing financial interests or personal relationships that could have appeared to influence the work reported in this paper

\section*{Acknowledgements}
The authors thank the PLaTIMed platform (\href{https://platimed.fr/}{https://platimed.fr/}) for accessing to the anatomical lab and helping in organising experiments.

\bibliography{main}

\end{document}